%% file: main.tex
\documentclass[runningheads]{llncs}
\usepackage{graphicx}

\usepackage{tikz}
\usepackage{comment}
\usepackage{amsmath,amssymb} 
\usepackage{color}
\usepackage[colorlinks]{hyperref}
\usepackage{booktabs}
\usepackage{subfigure}
\usepackage{blindtext}

\usepackage[accsupp]{axessibility}  

\usepackage{multirow}
\usepackage{wrapfig}

\begin{document}
\pagestyle{headings}
\mainmatter
\def\ECCVSubNumber{4931}  

\title{ARM: Any-Time Super-Resolution Method} 

\titlerunning{ARM: Any-Time Super-Resolution Method}
\authorrunning{Bohong Chen \emph{et al}.}
\author{Bohong Chen$^1$, Mingbao Lin$^{3}$, Kekai Sheng$^3$, Mengdan Zhang$^3$, \\Peixian Chen$^3$, Ke Li$^3$, Liujuan Cao$^{1}$\thanks{Corresponding Author}, Rongrong Ji$^{1,2,4}$}
\institute{
$^1$School of Informatics, Xiamen University. \\
$^2$Institute of Artificial Intelligence, Xiamen University. \\ $^3$Tencent Youtu Lab.~$^{4}$Institute of Energy Research, Jiangxi Academy of Sciences. \\
{\tt\small bhchen@stu.xmu.edu.cn, \quad linmb001@outlook.com}\\
{\tt\small \{saulsheng,davinazhang,peixianchen,tristanli\}@tencent.com,}\\
{\tt\small \{caoliujuan,rrji\}@xmu.edu.cn,}
}
\maketitle

\begin{abstract}
This paper proposes an Any-time super-Resolution Method (ARM) to tackle the over-parameterized single image super-resolution (SISR) models. Our ARM is motivated by three observations: 
(1) The performance of different image patches varies with SISR networks of different sizes. 
(2) There is a tradeoff between computation overhead and performance of the reconstructed image. 
(3) Given an input image, its edge information can be an effective option to estimate its PSNR.
Subsequently, we train an ARM supernet containing SISR subnets of different sizes to deal with image patches of various complexity.
To that effect, we construct an Edge-to-PSNR lookup table that maps the edge score of an image patch to the PSNR performance for each subnet, together with a set of computation costs for the subnets.
In the inference, the image patches are individually distributed to different subnets for a better computation-performance tradeoff.
Moreover, each SISR subnet shares weights of the ARM supernet, thus no extra parameters are introduced.
The setting of multiple subnets can well adapt the computational cost of SISR model to the dynamically available hardware resources, allowing the SISR task to be in service at any time.
Extensive experiments on resolution datasets of different sizes with popular SISR networks as backbones verify the effectiveness and the versatility of our ARM.
The source code is available at \url{https://github.com/chenbong/ARM-Net}.

\end{abstract}

\section{Introduction}
\label{sec:intro}

\begin{figure}[!t]
\centering
\includegraphics[width=0.96\textwidth]{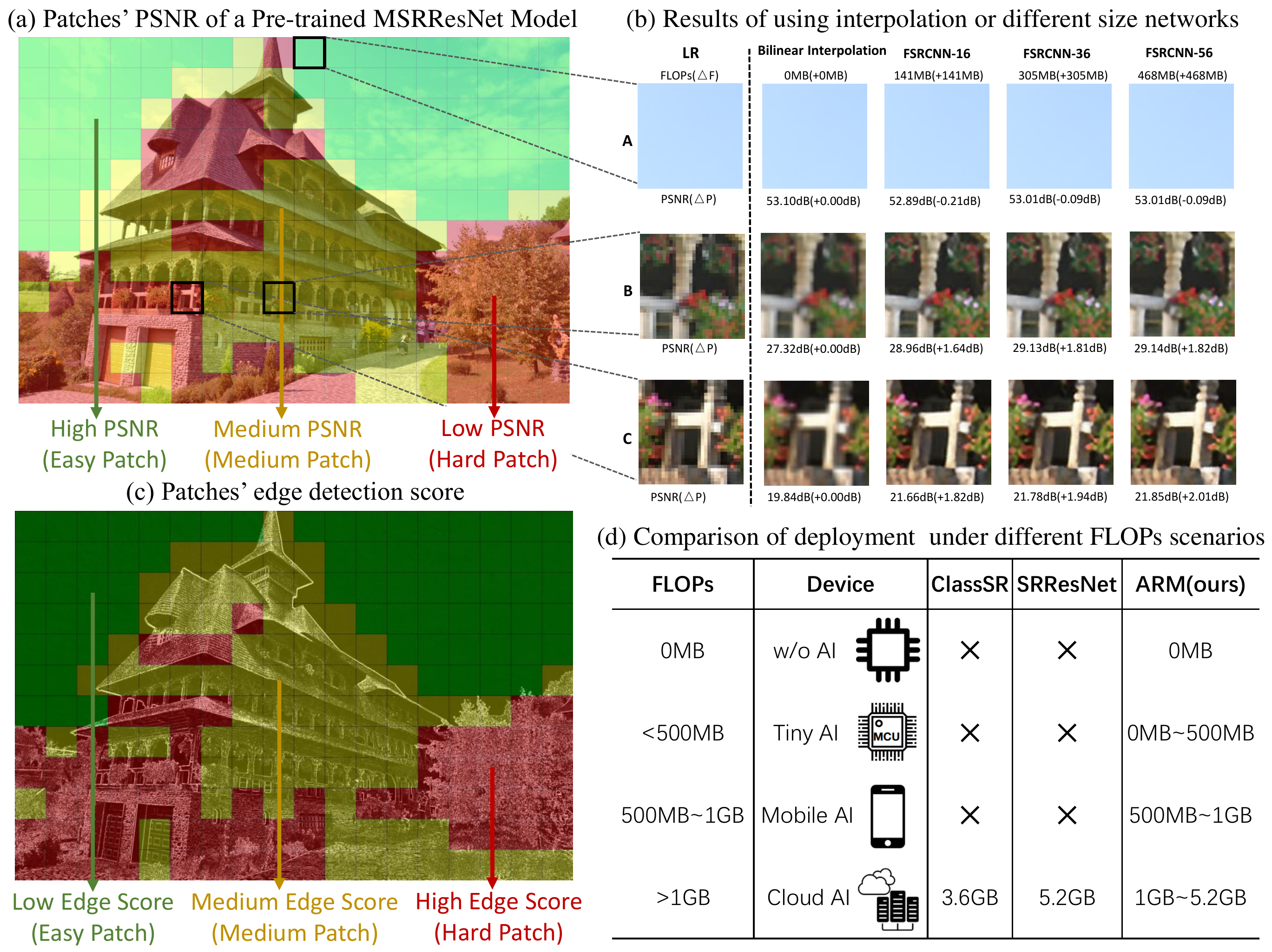}
\caption{Our observations in this paper. (a) Image patchs are categorized to three groups of ``easy''({\color{green}green}), ``moderate''({\color{yellow}yellow}) and ``difficult''({\color{red}red}) according to their PSNR via a pre-trained MSRResNet~\cite{wang2018recovering}. (b) Three groups prefer different super-resolution procedures: The ``easy'' patch benefits from the simple interpolation, the ``moderate'' patch benefits from a medium-sized SISR model, while ``difficult'' patch benefits from a large-sized SISR model. (c) Visualization of edge information. 
(d) Using SRResNet as a backbone, our ARM can support arbitrary size of FLOPs overhead without retraining.}
\label{fig:motivation}
\end{figure}

Recent years have witnessed the rising popularity of convolutional neural networks (CNNs) in the classic single image super-resolution (SISR) task that refers to constructing a high-resolution (HR) image from a given low-resolution (LR) version~\cite{Dong2016ImageSU,ledig2017photo,lim2017enhanced,yu2021path,zhan2021achieving,song2021addersr}.
SISR has a wide application in daily life such as facial recognition on low-resolution images, real-time video upscaling resolution on mobile devices, video quality enhancement on televisions, \emph{etc}.
For the sake of real-time experience, the SISR systems are required to be in service at any time.
However, the platforms to conduct SISR task are featured with: 
(1) The memory storage and computation ability are very limited. 
(2) The configured resources vary across different hardware devices. 
(3) The availability of hardware resources on the same device even changes greatly over different times.

Unfortunately, newly developed SISR models tend to have more learnable parameters as well as more floating-point operations (FLOPs). For example, the earliest CNN-based SISR model, SRCNN~\cite{dong2015image}, has only 3 convolutional layers with 57k parameters.
Later, VDSR~\cite{kim2016accurate} increases the number of parameters to 2.5M.
After that, RCAN~\cite{zhang2018image} 
increases its parameters to over 15M. 
where a total of 66,015G FLOPs are required to process one single 1,920$\times$1,080 image.
Consequently, existing SISR models can be barely deployed on the resource-hungry platforms. 
Therefore, how to design an efficient SISR network has attracted increasing interest in the computer vision community. Besides, how to dynamically adapt the SISR models to the currently available hardware resources for a real-time deployment also arouses the community's wide attention.
To this end, we investigate the computational redundancy in modern SISR networks and some observations are excavated as shown in Fig.\,\ref{fig:motivation}.

First, we observe that the performance of different image patches varies with SISR networks of different sizes.
In Fig.\,\ref{fig:motivation}(a), we categorize the image patches into three categories of ``easy" ({\color{green}green}),``moderate" ({\color{yellow}yellow}), and ``difficult" ({\color{red}red}) according to their PSNR scores from the pre-trained MSRResNet~\cite{wang2018recovering}.
Generally, we observe a higher PSNR for an ``easy'' patch, and vice versa. 
Then, we randomly pick up one patch from each of the three categories and super-resolution them up with the bilinear interpolation and FSRCNN network~\cite{dong2016accelerating} with different widths (\emph{i.e.}, the number of output channels) of 16/36/56. Visualization is shown in Fig.~\ref{fig:motivation}(b).
We find that for ``easy'' patch, the bilinear interpolation leads to the best results  while FSRCNN not only requires more computation, but degrades the performance.
In contrast, a wider FSRCNN benefits the ``hard'' patch.
This observation indicates that traditional SISR models suffer spatial redundancy from the input and it is necessary to deal with each patch based on its complexity for saving computation cost.

Second, for some image patches such as these from the ``moderate'' category, a medium-sized FSRCNN-36 can well complete the SR task while a larger FSRCNN-56 results in much heavier computation (+53\% FLOPs) with a very limited performance gain (+0.01 PSNR).
This observation indicates a tradeoff between computation overhead and performance of the reconstructed image. How to maintain the performance with a smaller computation burden deserves studying.
Finally, we find that ``hard'' patches often contain more edge information. To verify this, we perform edge detection~\cite{marr1980theory,canny1986computational} on each patch in Fig.~\ref{fig:motivation}~(c) and find a strong Spearman correlation coefficient between PSNR and negative edge score (see Fig.~\ref{fig:rank}~\textbf{Left}).
This observation implies that edge information can be an effective option to estimate PSNR of each patch with a cheaper computation cost since edge information can be obtained in an economical manner.

Inspired by the above observations, we propose an Any-time super-Resolution Method, referred to as ARM. 
Different from traditional CNNs-based SISR model with a fixed inference graph, our ARM dynamically selects different subnets for image super-resolution according to the complexity of input patches, and also dynamically adapts the computational overhead to the currently available hardware resources in inference.
Specifically, we first use a backbone network as the supernet, and train several weight-shared subnets of different sizes within the supernet. 
The weight-sharing mechanism avoids introducing extra parameters, leading to a light-weight SISR supernet.
Then, for each subnet, we construct an Edge-to-PSNR lookup table that maps the edge score of each patch to its estimated PSNR value.
Finally, 
we propose to choose the subnet for reconstructing a HR image patch with a larger output of PSNR prediction but a smaller computation cost to reach a computation-performance tradeoff.
We conduct extensive experiments on three large resolution datasets and results show that our ARM outperforms previous adaptive-based super-resolution ClassSR~\cite{kong2021classsr} with a computation-performance tradeoff. Moreover, comparing to ClassSR, with a better performance, our ARM reduces parameters by 78\%, 54\% and 52\% when using FSRCNN~\cite{dong2016accelerating}, CARN~\cite{ahn2018fast}, and SRResNet~\cite{ledig2017photo} as the supernet backbones.

Earlier studies~\cite{kong2021classsr,xie2021learning} also explore the spatial redundancy of image patches in SISR networks. They set up multiple independent branches to handle patches with different complexity. However, these operations increase the model parameters and can not adapt to the available hardware resources since the inference graph is static for a fixed image patch.
On the one hand, our ARM implements super-resolution in an economical manner since different image patches are fed to different small-size subnets of the supernet without introducing extra parameters.
On the other hand, our ARM enables a fast subnet switch for inference once the available hardware resources change. 
The main difference between the proposed ARM and the previous approach is shown in Fig.~\ref{fig:motivation}~(d), where the proposed ARM can theoretically support arbitrary computational overhead without retraining.
Thus the SISR task can be in service at any time.
%


\section{Related Work}

\noindent\textbf{Interpolation-based SISR}.
Image interpolation is built on the assumption of image continuity. Usually, it uses nearby pixels to estimate unknown pixels in high-resolution images by a fixed interpolation algorithm.
Classical interpolation algorithms include nearest neighbor interpolation, bilinear interpolation, bicubic interpolation and their variants~\cite{fekri1998generalized,thurnhofer1996edge}. The interpolation-based super-resolution benefits in cheap computation, but disadvantages in detail loss in the reconstructed images with complex textures.

\noindent\textbf{Region-irrelevant CNNs-based SISR}.
SRCNN~\cite{dong2015image} is the first work to build a three-layer convolutional network for high-resolution image reconstruction.
Since then, the successors~\cite{kim2016accurate,ledig2017photo,tong2017image} enhance the performance of SISR task by deepening the networks as well as introducing skip connections such as residual blocks and dense blocks.
EDSR~\cite{lim2017enhanced} reveals that the batch normalization (BN) layer destroys the scale information of the input images, which the super-resolution is sensitive to. Thus, it advocates removing BN layers in SISR task.
The increasing model complexity also arouses the community to design light-weight SISR models~\cite{dong2016accelerating,ahn2018fast,yu2018wide}. 
TPSR~\cite{lee2020journey} builds a network with the aid of network architecture search.
Zhan~\emph{et al.}~\cite{zhan2021achieving} combined NAS with parameter pruning to obtain a real-time SR model for mobile devices.
Many researches~\cite{li2020pams,ma2019efficient,xin2020binarized} are indicated to representing the full-precision SISR models with a lower-bit format. 
Also, the knowledge distillation is often considered to strengthen the quality of reconstructed images from light-weight models~\cite{gao2018image,he2020fakd,lee2020learning,wang2021towards}
Overall, these CNN models for SISR are often region-irrelevant, that is, the computational graph is never being adjusted to adapt to the input images.
On the contrary, our ARM picks up different subnets according to the complexity of input image patches, leading to a better tradeoff between model performance and computation.

\noindent\textbf{Region-aware CNNs-based SISR}.
Recently, several works have realized the redundancy of input images on spatial regions in the SISR task.
SMSR~\cite{wang2021exploring} introduces sparse masks upon spaces and channels within each block and then performs sparse convolution to reduce the computational burden.
FAD~\cite{xie2021learning} discovers that high-frequency regions contain more information and require a computationally intensive branch.
It adds multiple convolutional branches of different sizes and each feature map region is fed to one branch according to its region frequency.
ClassSR~\cite{kong2021classsr} classifies the image patches into three groups according to the patch difficulty. The image patches are respectively sent to the stand-alone trained backbone networks with different widths to realize super-resolution.

Despite the progress, the introductions of masks and additional network branches inevitably cause more parameters in these SISR models. Moreover, they fail to perform SISR task once the supportive hardware resources are insufficient.
On the contrary, our ARM innovates in its subnet weights sharing with the supernet. Thus, no additional parameters are introduced.
With multiple subnets, our SISR can well adapt to the hardware by a fast switch to perform the subnet capable of being run on the available resources.

\noindent\textbf{Resource-aware Supernet}.
In order to achieve dynamic adjustment of computational overhead during inference, many studies~\cite{yu2018slimmable,yu2019universally,yang2021mutualnet,lou2021dynamic} devise a supernet training paradigm with weight-shared subnets of multiple sizes. 
They uniformly sample different-size subnets for network training.
Compared with these resource-aware supernet training methods, our proposed ARM follows a similar supernet training paradigm but differs in the subnet selection that considers a tradeoff between computation and performance.

\section{Method}
\subsection{Preliminary}
Our ARM network is a CNN-based SISR supernet $\mathcal{N}_{W[0:1.0]}$ with its weights denoted as $W[0:1.0]$, where $1.0$ indicates a width multiplier of each convolutional layer. For example, a subnet can be represented as $\mathcal{N}_{W[0:0.5]}$ if its width of each convolutional layer is half of that in the supernet. 
By setting the width multiplier to different values, we can obtain a SISR subnet set $\{\mathcal{N}_{W[0:\alpha^j]}\}_{j=1}^M$ where $\alpha^j \in (0, 1]$ is the width multiplier of the $j$-th subnet and $M$ is the number of subnets.
Note that, $W[0:\alpha^j]$ is a subset of $W[0:1.0]$, \emph{i.e.}, $W[0:\alpha^j] \in W[0:1.0]$.
Therefore, each SISR subnet shares parts of weights in the SISR supernet.

\begin{figure}[!t]
\centering
\includegraphics[width=0.92\textwidth]{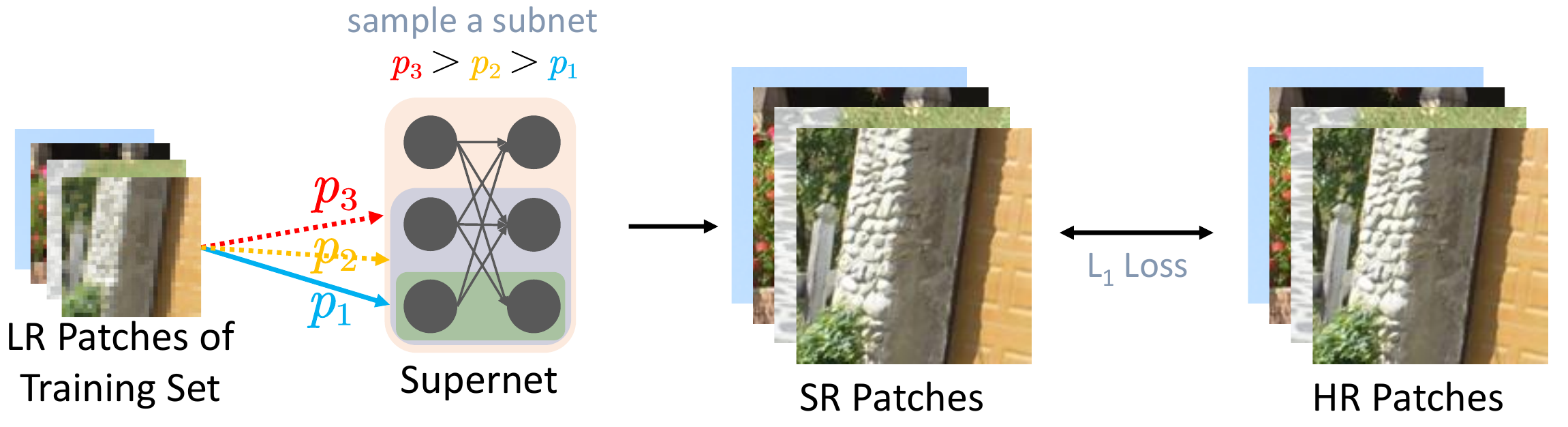}
\caption{Training of our ARM supernet. In each training iteration, one subnet is sampled to process a batch of low-resolution image patches with $\ell_1$ loss.}
\label{fig:training}
\end{figure}

For the traditional SISR task, a training set $\{X, Y\}$ can be available in advance, where $X=\{x_i\}_{i=1...N}$ are the low-resolution (LR) images, $Y=\{y_i\}_{i=1...N}$ are the high-resolution (HR) images, and $N$ is the number of training samples.
Our goal in this paper is to optimize the overall performance of each subnet with LR images $X$ as its inputs to fit the HR images $Y$. The formal optimization objective can be formulated as:
\begin{equation}\label{objective}
    \min_{W} \sum_{j=1}^M\frac{1}{N}\| \mathcal{N}_{W[0:\alpha^j]}(X) - Y\|_1,
\end{equation}
where $\|\cdot\|_1$ denotes the commonly-used $\ell_1$-norm in the SISR task\footnote{To stress the superiority of our method, we only consider $\ell_1$-norm loss in this paper. Other training losses~\cite{fuoli2021fourier,johnson2016perceptual} for SISR can be combined to further enhance the results.}.
Note that, to update the subnets is to update the supernet in essence since weights of each subnet are a subset of the supernet weights. 
Any existing SISR network can serve as the backbone of our supernet $\mathcal{N}_{W[0:1.0]}$. Fig.\,\ref{fig:training} illustrates the training paradigm of our ARM supernet for an any-time SISR task. 
More details are given in the following context.

\subsection{ARM Supernet Training}\label{sec:supernet_training}
The optimization of Eq.\,(\ref{objective}) requires all subnets to involve in the network training, which causes a heavy training consumption. 
Instead, we propose to iteratively optimize each subnet based on a divide-and-conquer manner by performing optimization for a particular subnet $\mathcal{N}_{W[0:\alpha^j]}$ in each iteration. In this case, the suboptimization problem becomes:
\begin{equation}\label{subobjective}
    \min_{W}\frac{1}{N'}\| \mathcal{N}_{W[0:\alpha^j]}(X') - Y' \|_1,
\end{equation}
where $X' = \{x'_i\}_{i=1...N'} \in X$ and $Y' = \{y_i\}_{i=1...N} \in Y$ are respectively the LR image batch and HR image batch in the current training iteration\footnote{In our supernet training, $X'$ and $Y'$ are indeed batches of local image patches from the $X$ and $Y$. Details are given in Sec.\,\ref{settings}. For brevity, herein we simply regard them as image batches.}.
The rationale behind our divide-and-conquer optimization is that the overall objective is minimized if and only if each stand-alone subnet in Eq.(\ref{objective}) is well optimized. Thus, we choose to decouple the overall objective and optimize individual subnets respectively, which also leads to the minimization of Eq.(\ref{objective}).

So far, the core of our supernet training becomes how to choose a subnet for optimization in each training iteration. 
Traditional supernet training in image classification tasks~\cite{yu2018slimmable,yu2019universally} usually adopts the uniform sampling where the subnets of different sizes are endowed with the same sampling probability.
Though applicable well in the high-level classification, we find the sampling strategy could impair SISR task (see the numerical values in Sec.\,\ref{sec:sampling}).
Empirically, in the supernet training: i) smaller subnets tend to overfit the inputs and require less training, ii) larger subnets tend to underfit the inputs, therefore require more training.
Consequently, we believe that the performance impairment of uniform subnet sampling is due to the different degree of training required by different size subnets.

Therefore, we propose a computation-aware subnet sampling method where subnets with heavier computation burden are endowed with a higher training priority. Concretely, the sampling probability of the $j$-th subnet is defined as:
\begin{equation}\label{probability}
p^j = (\frac{FLOPs(\mathcal{N} _{W[0:\alpha^j]})^2)}{\sum_{k=1}^M{FLOPs(\mathcal{N} _{W[0:\alpha^k]})^2)}},
\end{equation}
where $FLOPs(\cdot)$ calculates the FLOPs consumption of its input subnet.

To sum up, in the forward propagation, we sample a subnet with its probability of Eq.\,(\ref{probability}) to perform loss optimization of Eq.\,(\ref{subobjective}). In the backward propagation, we only update weights of the sampled subnet.
Consequently, after training, we obtain a supernet $\mathcal{N}_{[0:1.0]}$ containing $M$ high-performing subnets $\{\mathcal{N}_{W[0:\alpha^j]}\}_{j=1}^M$.
It is a remarkable fact that our ARM supernet does not introduce additional parameterized modules for its weight-sharing mechanism. 
In addition to the ARM supernet, we have added an interpolation branch, \emph{i.e.} using interpolation directly for super-resolution, denoted as a special subnet $\mathcal{N}_{W[0:\alpha^0]}$, where $\alpha^0=0$.
This interpolation branch can be run on the device with very low computational performance, \emph{i.e.}, the ``w/o AI" device in Fig.\ref{fig:motivation}~(d).
Thus the $M+1$ subnets in our ARM supernet can be further expressed uniformly as $\{\mathcal{N}_{W[0:\alpha^j]}\}_{j=0}^M$.
Besides, our ARM eliminates the spatial redundancy of the inputs by adapting the complexity of image patches to different subnets for a computationally economical inference, as detailed below.

\begin{figure}[!t]
\centering
\includegraphics[width=0.95\textwidth]{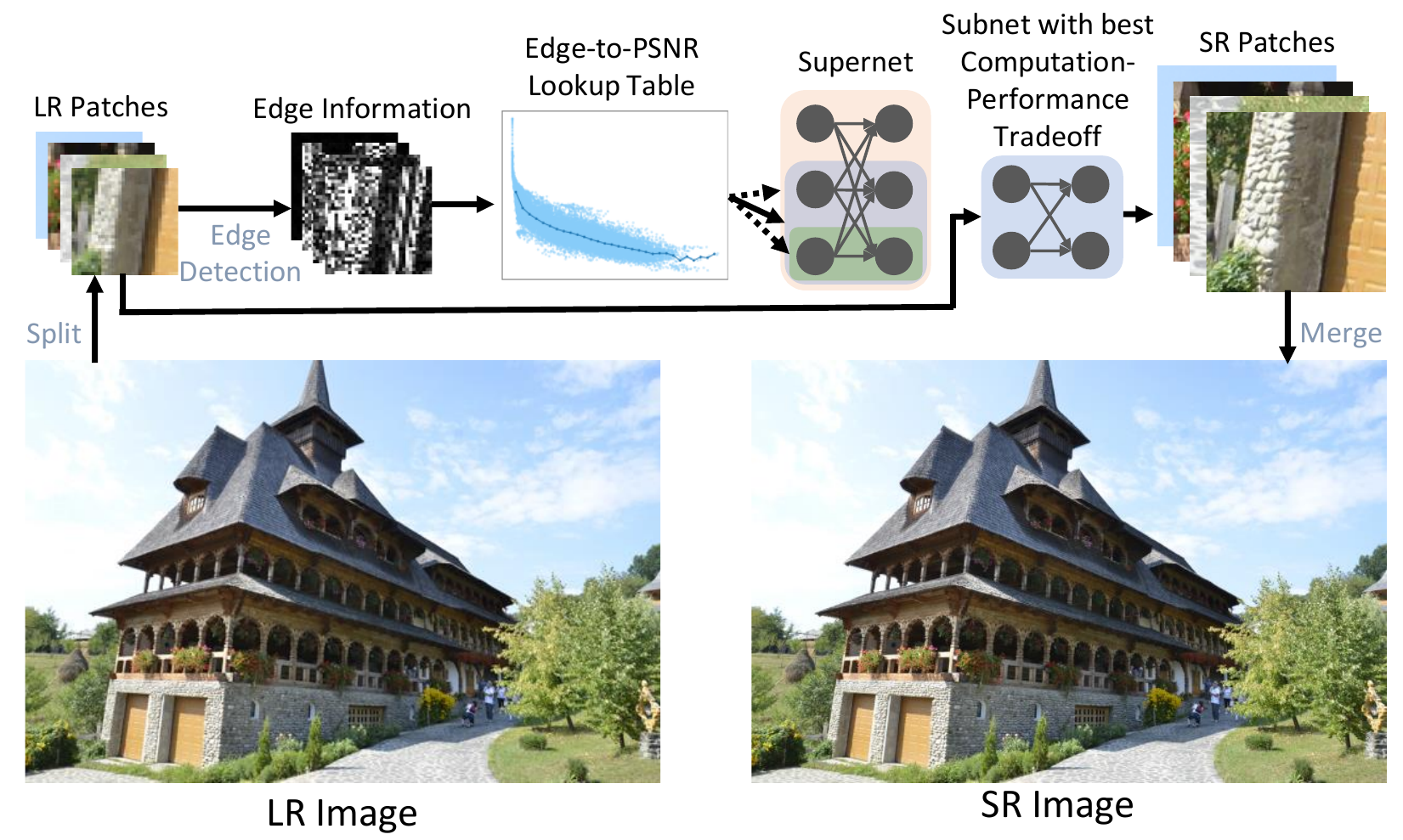}
\caption{Inference of our ARM supernet. We first detect the edge information of the local LR image patches, PSNR performance of which is then estimated by the pre-built Edge-to-PSNR lookup tables. The subnet with a best computation-performance tradeoff is then selected to construct HR version of the input LR image patches.}
\label{fig:infer}
\end{figure}

\subsection{ARM Supernet Inference}
Fig.\,\ref{fig:infer} outlines the pipeline of our ARM supernet inference.
In short, we first split a complete LR image into several local patches of the same size, and performs the edge detection on these LR patches to calculate their edge scores.
Then, the PSNR performance of each patch is estimated by our pre-built Edge-to-PSNR lookup table.
For each subnet, we further pre-calculate its computation cost, and propose to choose the subnet for SR inference with a larger PSNR output but a smaller computation cost in order to pursue a computation-performance tradeoff.
Finally, the SR patches are merged to recover the complete SR image.

\begin{figure}[t]
\centering
\begin{minipage}[t]{0.4\textwidth}
\centering
\includegraphics[width=0.8\textwidth]{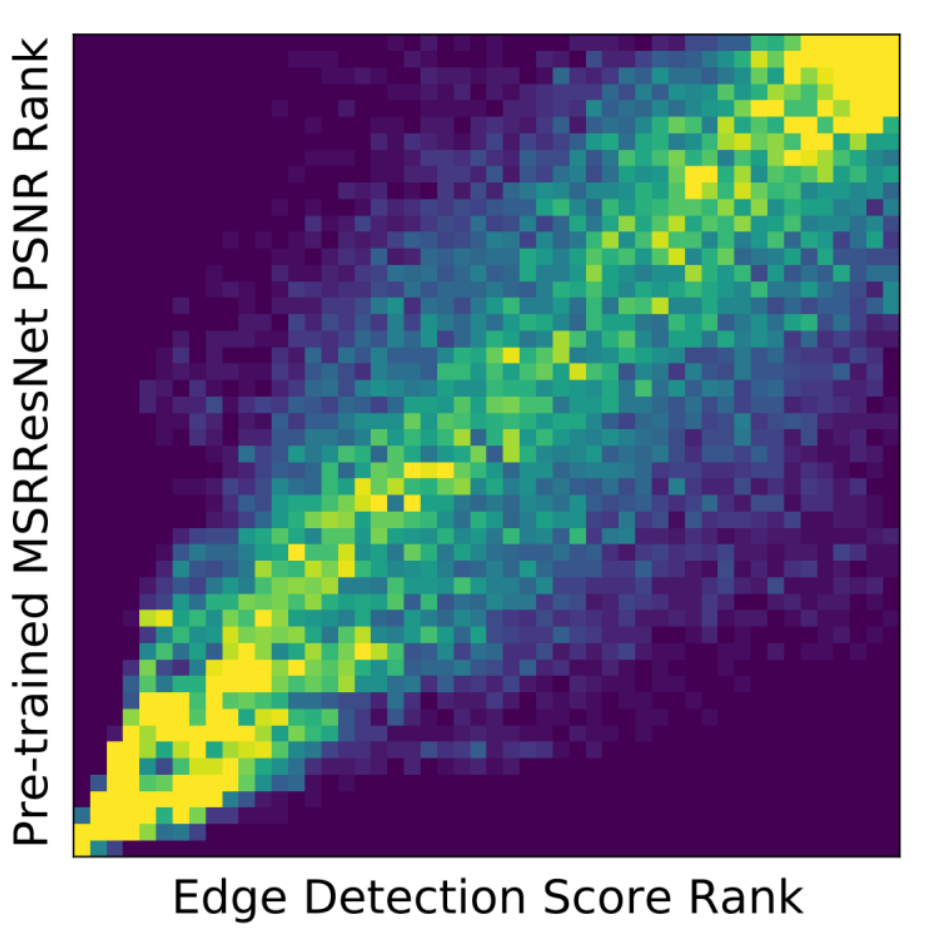}
\end{minipage}
\begin{minipage}[t]{0.55\textwidth}
\centering
\includegraphics[width=1\textwidth]{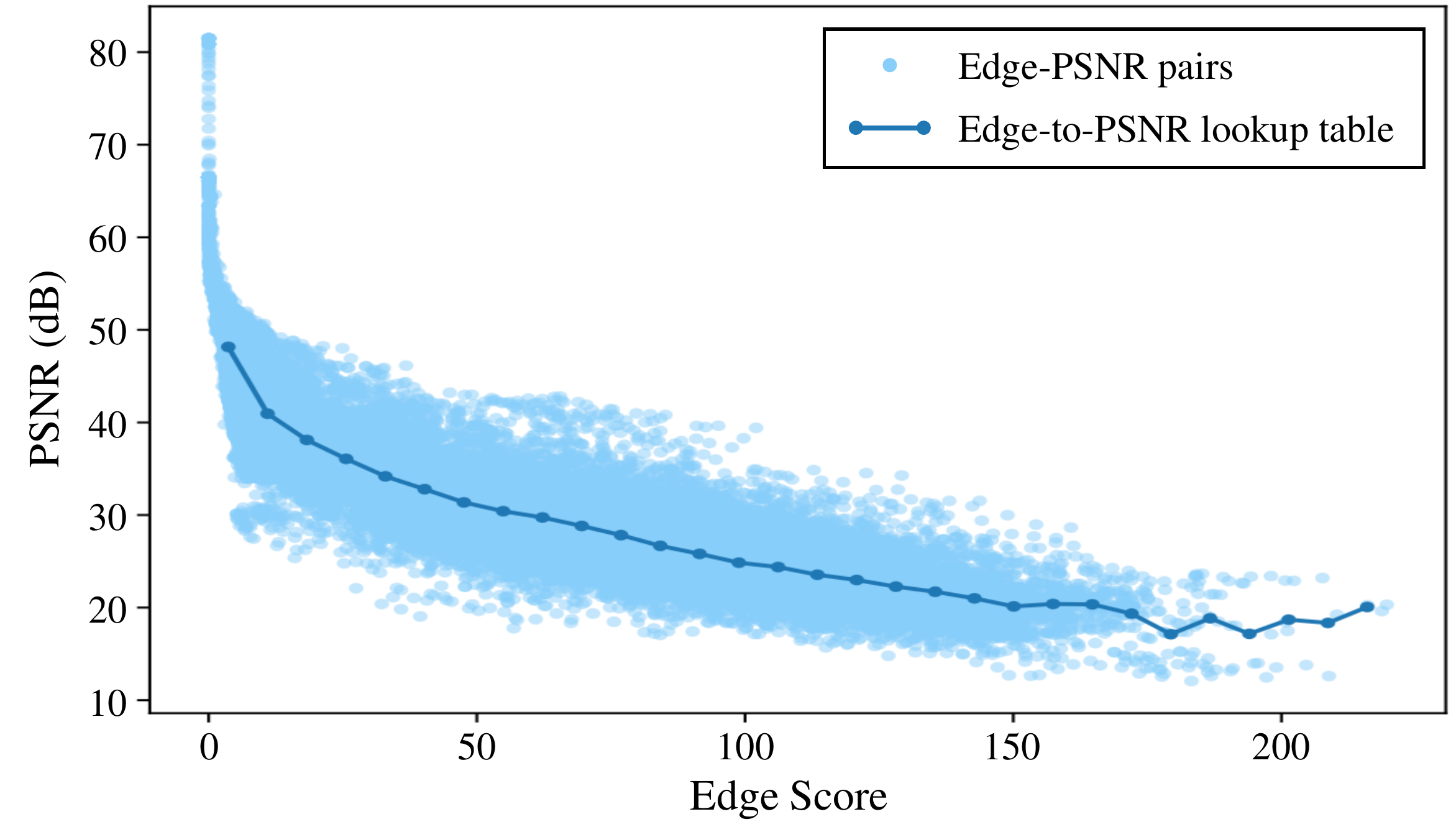}
\end{minipage}
\caption{
\textbf{Left:} A strong Spearman correlation coefficient (0.85) between PSNR and negative edge score.\label{fig:rank}
\textbf{Right:} Edge-PSNR pairs and the Edge-to-PSNR lookup table.\label{fig:train_3}
}
\end{figure}

\noindent\textbf{Edge Score}.
\label{sec:edge_score}
Recall that in Sec.\,\ref{sec:intro}, we observe a strong correlation between the image edge and the PSNR performance, which is also verified in Fig.\,\ref{fig:rank}~\textbf{Left}. To measure the edge information, we first generate the edge patches using the laplacian edge detection operator~\cite{marr1980theory} that allows observing the features of a patch for a significant change in the gray level. A large pixel value indicates richer edge information. Thus, we define the mean value of all pixels in a gray edge patch as the edge score to reflect the overall information in an edge patch.

\noindent\textbf{Edge-to-PSNR Lookup Table}.
\label{sec:lookup_table}
In contrast to a heavy network inference to derive the PSNR of each patch, the edge-psnr correlation inspires us to construct a set of Edge-to-PSNR lookup tables $T=\{ t^j \}_{j=0}^M$ where the $j$-th lookup table $t^j$ maps the patch's the edge score to the estimated PSNR performance for the corresponding subnet $\mathcal{N}_{W[0:\alpha^j]}$.

To be more specific, we first compute the edge score set for all the LR patches from the validation set of DIV2K~\cite{Agustsson2017NTIRE2C}, denoted as $E = \{e_i\}_{i=1}^O$ where $e_i$ is the edge score of the $i$-th patch and $O$ is the patch size.
We split the edge score interval $[\min(E), \max(E)$] equally into a total of $K$ subintervals $S = \{s_k\}_{k=1}^K$, where $\min(\cdot)$ and $\max(\cdot)$ return the minimum and maximum of the input edge score set. As a result, we know that the scope of the $k$-th subinterval falls into:
\begin{equation*}
\footnotesize{
    s_k = [ \min(E) + \frac{\max(E)-\min(E)}{K}\cdot (k-1), \;
            \min(E) + \frac{\max(E)-\min(E)}{K}\cdot k].
}
\label{subinterval}
\end{equation*}

After the supernet training, these LR patches are further fed to the trained subnet $\mathcal{N}_{W[0:\alpha^j]}$ to obtain their reconstructed HR patches.
Then, the PSNR values between SR patches and HR patches are computed, denoted as $P^j = \{ p_i^j \}_{i=1}^O$.
Based on the subinterval splittings, we calculate the average PSNR within the $k$-th subinterval for the subnet $\mathcal{N}_{W[0:\alpha^j]}$ as:
\begin{equation}\label{average}
    \bar{p}_k^j = \frac{\sum_{i=1}^O \mathcal{I}(e_i \in s_k) \cdot p_i^j}{\sum_{i=1}^O\mathcal{I}(e_i \in s_k)},
\end{equation}
where $\mathcal{I}(\cdot)$ is an indicator which returns 1 if the input is true, and 0 otherwise.

Subsequently, our $j$-th Edge-to-PSNR lookup table $t^j$ is defined as: $S \rightarrow \bar{P}^j = \{\bar{p}_k^j\}_{k=1}^K$.
For an illustrative example, we use FSRCNN~\cite{dong2016accelerating} as our supernet backbone and three subnets with $\alpha_1 = 0.29$, $\alpha_2 = 0.64$ and $\alpha_3 = 1.0$ are trained on DIV2K dataset~\cite{Agustsson2017NTIRE2C}. In Fig.~\ref{fig:train_3}~\textbf{Right}, we plot the statistical results of the first subnet $\mathcal{N}_{W_{[0:\alpha_1]}}$ including edge-psnr pairs of $\{(e_i, p_i^1)\}_{i=1}^O$ and Edge-to-PSNR lookup table $t_1$. Generally, our lookup table can well fit the distribution of edge-psnr pairs, thus it can be used to estimate the PSNR performance.

\noindent\textbf{Computation-Performance Tradeoff}.
\label{sec:computation_tradeoff}
Given a new LR image patch during inference, we derive its edge score $\hat{e}$ first, and it is easy to know that this patch falls into the $\hat{k}$-th subinterval where $\hat{k} = \lfloor\frac{(\hat{e}-\min(E))\cdot K}{\max(E)-\min(E)} + 1\rfloor$, in which $\lfloor\cdot\rfloor$ is a floor function. 
With the pre-built lookup tables, we can easily obtain the estimated PSNR $\bar{p}^j_{\hat{k}}$ for the subnet $\mathcal{N}_{W[0:\alpha^j]}$.
It is natural to choose the subnet with the best predicted PSNR to reconstruct the HR version of a given LR image patch. In this case, the selected subnet index is:
$\mathop{\arg\max}_j  \;\bar{p}^j_{\hat{k}}$.

However, our observation in Sec.\,\ref{sec:intro} indicates a tradeoff between the computation and performance. A larger subnet may result in a slightly better performance, but a much heavier increase in computation. 
Given this, we further propose to take into consideration the computation burden of each subnet.
To that effect, we further maintain a set of computation costs $C = \{c^j\}_{j=0}^M$ where $c^j$ denotes the computation of the $j$-th subnet $\mathcal{N}_{W[0:\alpha_j]}$.
Then, we propose a computation-performance tradeoff function based on the estimated PSNR performance as well as the computation cost $c^j$ to pick up a subnet for SISR as:
\begin{equation}
\label{eq:select}
    \hat{j} = \mathop{\arg\max}\limits_j \; \eta \cdot \bar{p}^j_{\hat{k}} - c^j,
\end{equation}%
where $\eta$ is a hyper-parameter to balance the numerical difference between computation cost and PSNR estimation during inference stage. Its influence will be studied in supplementary materials. As result, the subnet $\mathcal{N}_{W_{[0:\alpha_{\hat{j}}]}}$ is used to deal with the input patch.
And we accomplish the goal to deal with each image patch according to its complexity, as observed in Sec.\,\ref{sec:intro}.

We would like to stress that the lookup tables $T = \{t^j\}_{j=0}^M$ and the computation cost set $C=\{c^j\}_{j=0}^M$ are built offline and once-for-all. 
In the online inference stage, the PSNR values of new image patches can be quickly fetched from the pre-built lookup tables according to their edge information which also can be quickly derived by off-the-shelf efficient edge detection operator~\cite{marr1980theory,canny1986computational}.
Thus, our ARM supernet does not increase any computation burden for the SISR task. Moreover, the setting of multiple subnets within a supernet enables a fast subnet switch for inference once the available hardware resources change. Therefore, our ARM can be in service at any time.

\section{Experiments}
\subsection{Settings}\label{settings}

\noindent\textbf{Training}.
Without loss of generality, we construct the ARM supernets on three typical types of SISR backbones of different sizes, including FSRCNN-56~\cite{dong2016accelerating} (small), CARN-64~\cite{ahn2018fast} (medium), and SRResNet-64~\cite{ledig2017photo} (large), where 56, 64, and 64 are the width of the corresponding supernets.
We set three subnets in each supernet, and the width multipliers ($\alpha_j$) of them are (0.29, 0.46, 1.0) for FSRCNN, (0.56, 0.81, 1.0) for CARN and SRResNet, which are the same model configurations of ClassSR~\cite{kong2021classsr} for fair comparison.
Moreover, the images of the training set will be pre-processed into small patches in the same way as ClassSR~\cite{kong2021classsr} before our ARM supernet training. Also, the training set of DIV2K (index 0001-0800)~\cite{Agustsson2017NTIRE2C} is used to train our ARM supernet. 
During training, the data augmentation includes random rotation and random flipping.
The training process lasts 20 epochs with a batch size of 16.
The ADAM optimizer~\cite{kingma2015adam} is applied
with $\beta_1=0.9$ and $\beta_2=0.999$. 
The training costs for FSRCNN, CARN and SRResNet are 4, 22 and 24 GPU hours on a single NVIDIA V100 GPU, respectively.
For more details, please refer to \href{https://github.com/chenbong/ARM-Net}{our code}.

\input{tab_MRonFlickr2KTest2K}

\noindent\textbf{Benchmark and Evaluation Metric}.
We apply the Peak signal-to-noise ratio (PSNR) as the metric to evaluate SR performance of all methods on four test sets: F2K, Test2K, Test4K, and Test8K.
F2K consists of the first 100 images (index 000001-000100) of Flickr2K~\cite{lim2017enhanced} dataset. Test2K, Test4K, and Test8K datasets are constructed from DIV8K~\cite{gu2019div8k} following the previous work~\cite{kong2021classsr}. Other evaluation settings are also the same as the standard protocols in~\cite{kong2021classsr}.
Unless otherwise stated, the FLOPs in this paper are calculated as the average FLOPs for all 32$\times$32 LR patches with $\times$4 super-resolution across the whole test set.

\input{tab_MRonTest4KTest8K}

\subsection{Main Results}
\begin{wrapfigure}{r}{0.43\textwidth}
\centering
\includegraphics[width=0.43\textwidth]{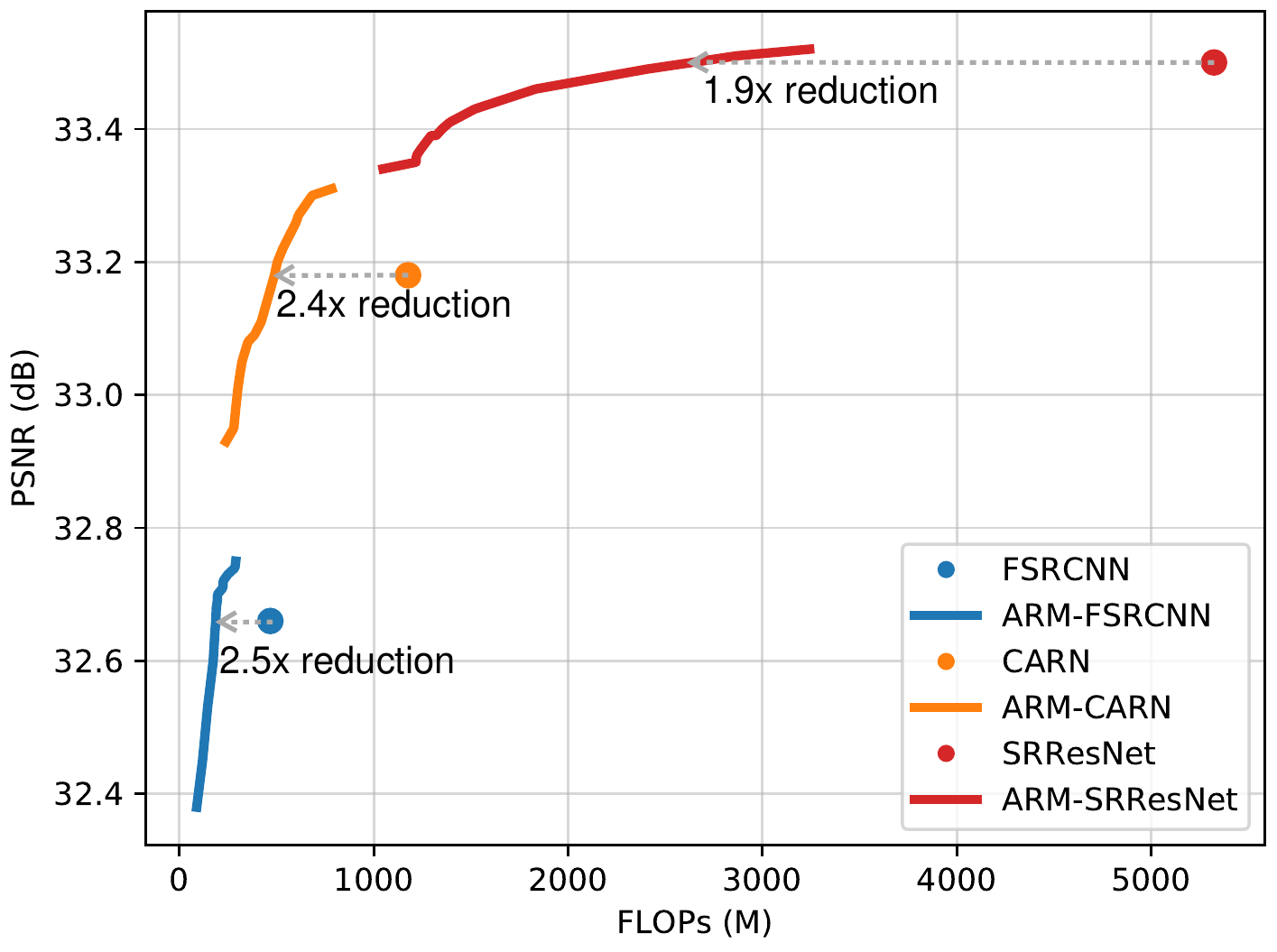}
\caption{ARM networks use the same parameters as the backbones, but offer adjustable computations and thus better FLOPs-PSNR tadeoffs.}
\label{fig:psnr_flops}
\end{wrapfigure}

\begin{figure}[t]
\centering
\includegraphics[width=\textwidth]{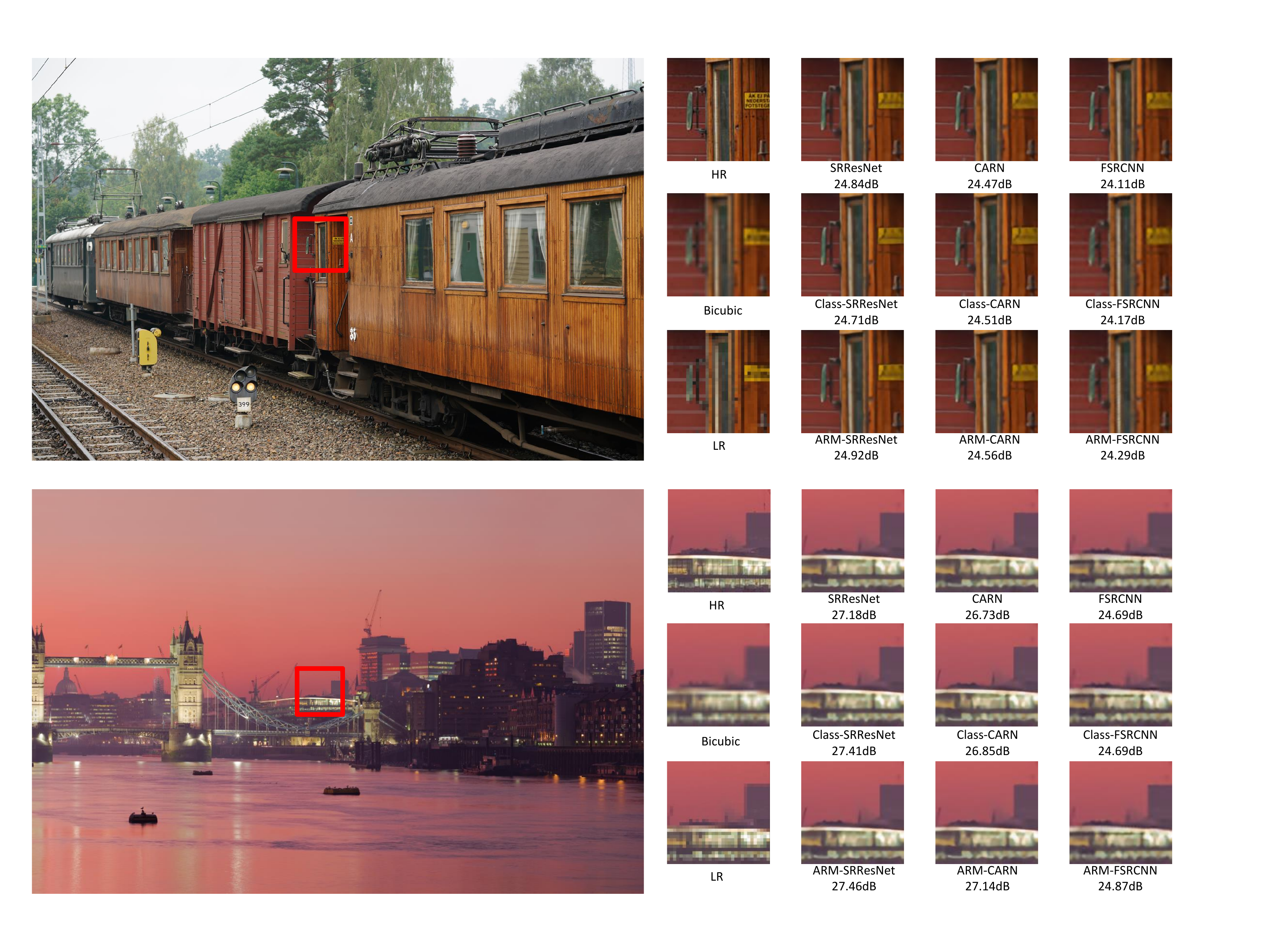}
\caption{Quantitative comparison of ARM networks with backbone networks, and SOTA dynamic SISR method~\cite{kong2021classsr} with $\times$4 super-resolution. These two examples are image ``1294"(above) from Test2K and image ``1321"(below) from Test4K, respectively. ARM produce a higher PSNR compared to backbone networks and SOTA methods.}
\label{fig:compare}
\end{figure}

As listed in Table~\ref{tab:MRonFlickr2KTest2K} and Table~\ref{tab:MRonTest4KTest8K}, our ARM network achieves better results with less computation than the three SISR backbones.
Overall, as we can observe in Fig.~\ref{fig:psnr_flops}: compared to the backbones, the width of the ARM network can be dynamically adjusted to achieve the computation-performance tradeoff.
Since our method is the most similar to ClassSR, for fair comparisons, our experimental settings are basically the same as ClassSR.
The results verify that ARM achieves comparable or better results than ClassSR in most cases.
On the three backbones, ARM outperforms ClassSR in terms of computation-performance tradeoff on both the Test2K and Test8K datasets.
Fig.~\ref{fig:compare} also indicate that our ARM can better reconstruct the HR images.

\subsection{Computation Cost Analysis}
\label{sec:abexp_cost}
\textit{In the training phase}, ARM differs from a normal backbone network only in the way the subnets of each batch are selected. Therefore, there is no additional computational and parameter storage overhead for training ARM networks. However, ClassSR has more parameters to be updated due to the higher number of parameters, thus incurring additional computational and parameter storage overheads.
After the training, ARM needs to use the validation set of DIV2K (index 801-900) for inference to construct the Edge-to-PSNR lookup tables $T$, this step will incur additional inference overhead and CPT function storage parameters storage overhead.
Luckily, since only forward and not backward gradient updates are required, this step is quick, taking only a few minutes, and it can be done offline only once. 
Actually, the Edge-to-PSNR lookup tables $T$ take only $M\times K$ parameters, where $M$ is the number of subnets of different widths and $K$ is the number of subintervals (
see Sec.~\ref{sec:supernet_training}). In this paper, we take $K=30$. Thus for $M=3$ subnets, only $90$ additional parameters are needed, which is almost negligible compared to the large amount of parameters in the original network.

\textit{In the inference}, the input patch needs to be edge detected. The FLOPs for edge detection are about 0.02M, which is also almost negligible compared to 468M FLOPs for FSRCNN, 1.15G FLOPs for CARN and 5.20G FLOPs for SRResNet. Thus, our inference is very efficient.

\subsection{Ablation Study}
\label{ablation}

\begin{figure}[t]
\centering
\begin{minipage}[t]{0.32\textwidth}
\centering
\includegraphics[width=1\textwidth]{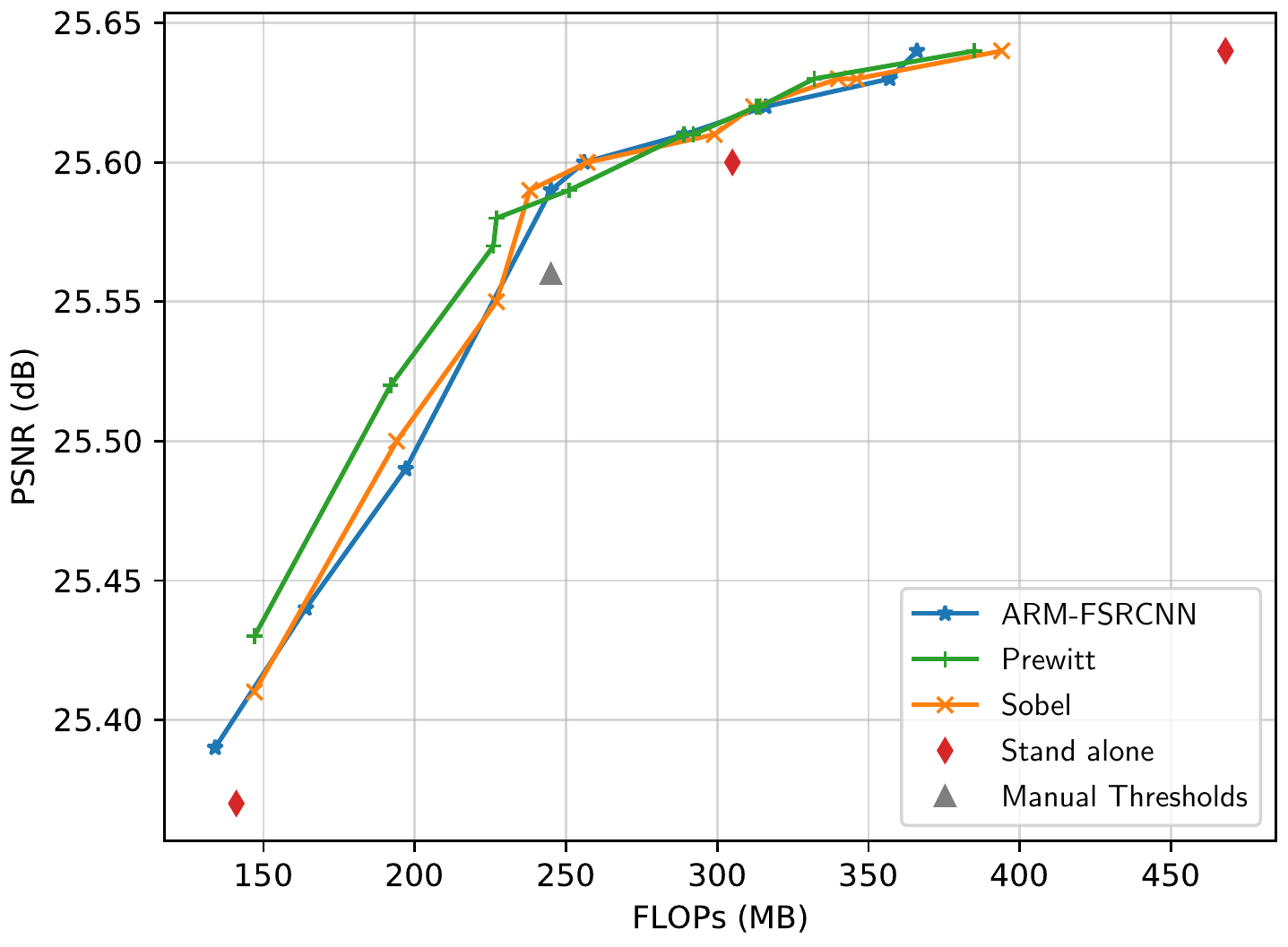}
\end{minipage}
\begin{minipage}[t]{0.32\textwidth}
\centering
\includegraphics[width=1\textwidth]{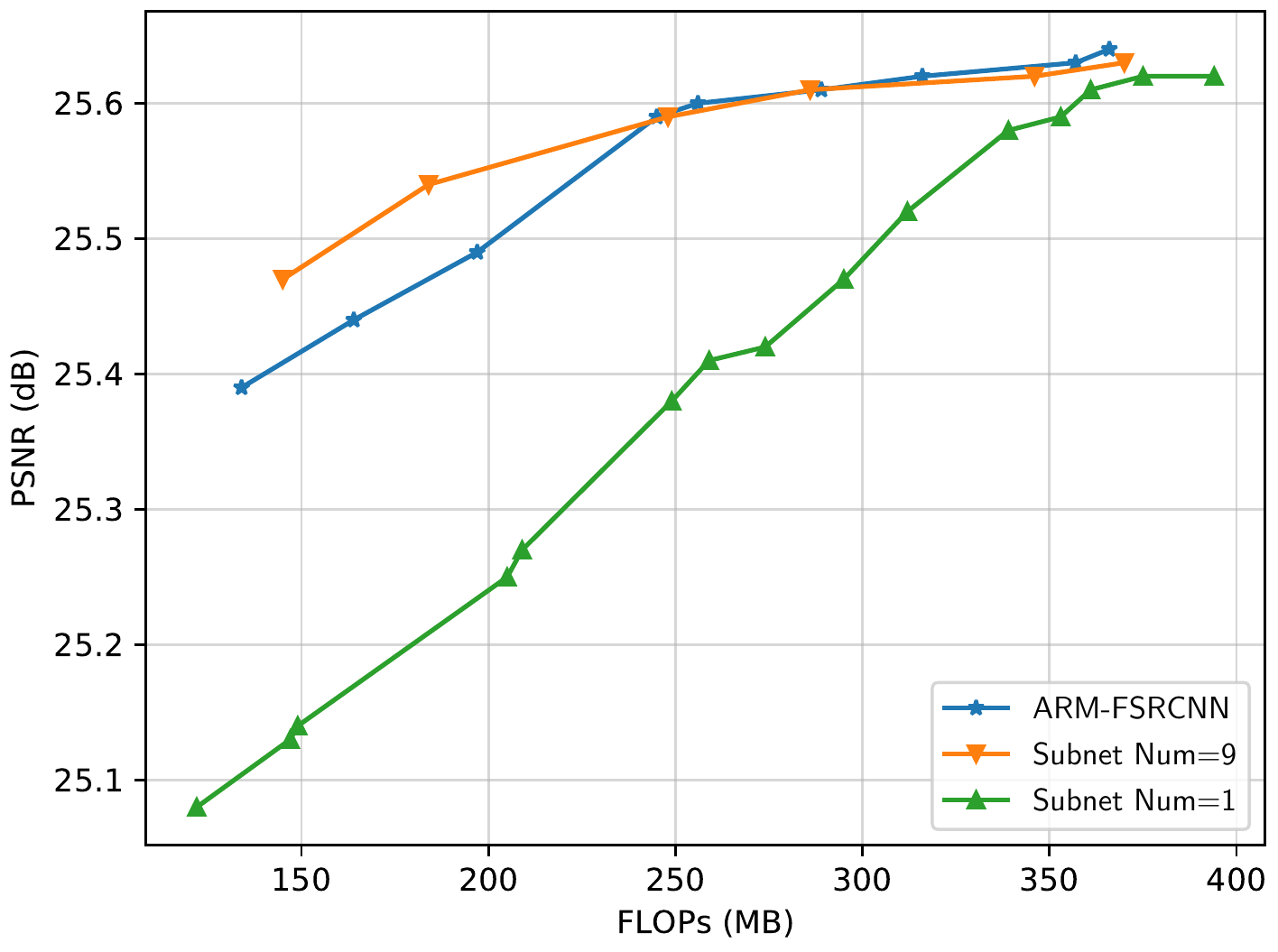}
\end{minipage}
\begin{minipage}[t]{0.32\textwidth}
\centering
\includegraphics[width=1\textwidth]{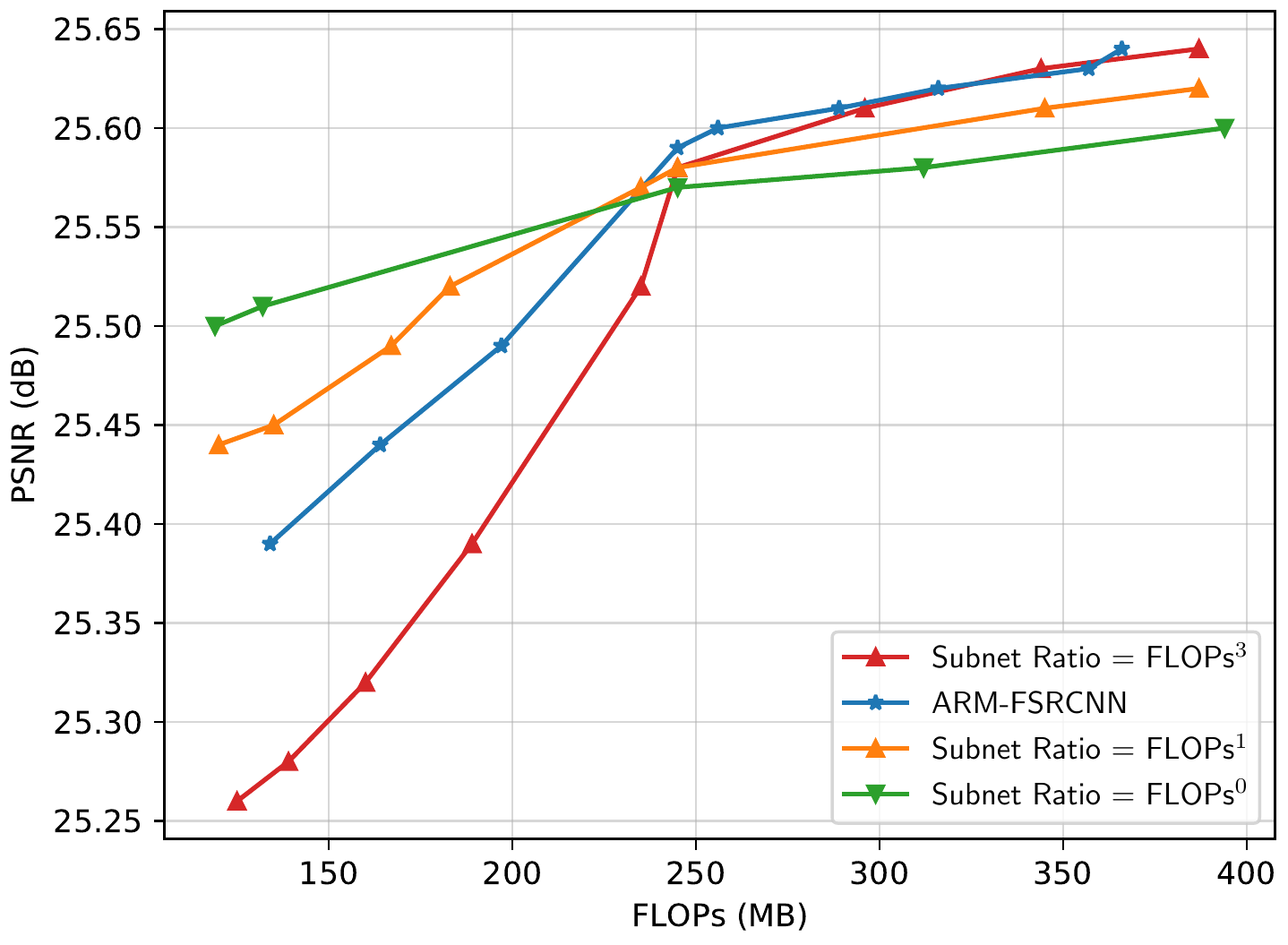}
\end{minipage}
\caption{
\textbf{Left:} Effectiveness of lookup tables and several edge detection operators.\label{fig:ab1}
\textbf{Middle:} Different numbers of subnets.\label{fig:ab2}
\textbf{Right:} Various subnet sampling ratios.\label{fig:ab3}
}
\end{figure}

\noindent\textbf{Edge-to-PSNR Lookup Tables.}
To verify the effectiveness of subnet selection via the Edge-to-PSNR lookup tables, we compared the computation-performance tradeoff of different strategies on the same pre-trained ARM-FSRCNN supernet: using the Edge-to-PSNR lookup tables, manually setting the edge score threshold, and using each size of subnet.
For an ARM network with one interpolation branch and three different sizes networks, each patch has four choices and needs to be classified into four categories according to the edge score during inference. There are many optional strategies to divide the patches. For simplicity, we apply the most intuitive one by calculating the edge scores of all patches on the validation set of DIV2K and sorting them to obtain thresholds that allow the validation set to be averaged into four categories. The 
thresholds are then used in the test set.
We also test the performance of different width subnets of the ARM supernet.
As illustrated in Fig.~\ref{fig:ab1}~\textbf{Left}, the subnet selection with the Edge-to-PSNR lookup tables outperforms 
both the manual setting of thresholds ({\color{gray}$\blacktriangle$} in Fig.~\ref{fig:ab1} \textbf{Left}) and the three separate subnets ({\color{red}$\blacklozenge$} in Fig.~\ref{fig:ab1}~\textbf{Left}).
The results ensure that using the Edge-to-PSNR lookup tables is indeed more effective in selecting subnets according to the specificity of the patch.

\noindent\textbf{Edge Detection Operators.}
\label{sec:edge_operator}
Recall that in Sec.~\ref{sec:edge_score}, ARM use the edge detection operator to obtain the edge score of patches, and ARM uses the laplacian as the edge detection operator by default. Here, we conduct ablation experiments on the types of edge detection operators.
Specifically, besides the default laplacian operator, we also tried the Sobel and Prewitt operators. The results are shown as Fig.~\ref{fig:ab1}.
Different operators achieve good FLOPs-PSNR tradeoff, and it indicates that our ARM is robust to different edge detection operators.

\noindent\textbf{The Number of Subnets.}
\label{sec:subnet_number}
As shown in Fig.~\ref{fig:ab2}~\textbf{Middle}, a larger number of subnets results in a better tradeoff, especially under lower computation resources. Hence, setting the number of subnets to $3$ is not optimal.
However, for a fair comparison with ClassSR, we still set the number of subnets to $3$.

\noindent\textbf{Subnet Sampling.}
\label{sec:sampling}
As we point out in Sec.~\ref{sec:supernet_training}, uniform sampling can lead to performance degradation.
In the main experiment, the sampling probabilities of different sized subnets are set to be proportional to the $n$-th power of subnet FLOPs, which can be denoted as FLOPs$^n$: $p^j = (\frac{FLOPs(\mathcal{N} _{W[0:\alpha^j]})^n)}{\sum_{k=1}^M{FLOPs(\mathcal{N} _{W[0:\alpha^k]})^n)}}$, where $M$ is the number of subnets of different widths.
The sampling probabilities of different sized subnets of ARM are set to FLOPs$^2$ by default.
We experiment with different sampling ratios such as FLOPs$^0$, \emph{a.k.a.} uniform sampling, FLOPs$^1$, FLOPs$^2$ and so on. The results are shown in Fig.~\ref{fig:ab3}~\textbf{Right}.
When using uniform sampling, the smaller networks are entitled with enough chances to be selected for training.
Then the performance under smaller FLOPs will be better, but the performance at larger FLOPs is degraded.
As the sampling probability of large subnets gradually increases, \emph{e.g.}, FLOPs$^1$, FLOPs$^2$ and FLOPs$^3$ in Fig.~\ref{fig:ab3}~\textbf{Right}, the performance at large FLOPs gradually improves.
On the contrary, the performance at small FLOPs gradually decreases.
In our experiments, we focus more on the performance of the ARM at larger FLOPs. As shown in Fig.~\ref{fig:ab3}~\textbf{Right}, FLOPs$^3$ has limited performance improvement over FLOPs$^2$ on large FLOPs, but much lower performance on small FLOPs. Thus, in this paper, we set the sampling probability of the subnets to FLOPs$^2$ by default.

\section{Conclusion}
In this paper, we introduce an ARM supernet method towards any-time super-resolution. Several weight-shared subnets are trained separately to deal with images patches of different complexities for saving computation cost.
We observe that the edge information can be an effective option to estimate the PSNR performance of an image patch. Subsequently, we build an Edge-to-PSNR lookup table for each subnet to pursue a fast performance estimation. On the basis of lookup tables, we further propose a computation-performance tradeoff function to pick up a subnet for constructing a HR version of the given LR image patch. This leads to a supreme performance of our ARM in SISR task, as well as significant reduction on computation cost.\\

\noindent\textbf{Acknowledgments.} This work was supported by the National Science Fund for Distinguished Young Scholars (No.62025603), the National Natural Science Foundation of China (No. U21B2037, No. 62176222, No. 62176223, No. 62176226, No. 62072386, No. 62072387, No. 62072389, and No. 62002305), Guangdong Basic and Applied Basic Research Foundation (No.2019B1515120049), and the Natural Science Foundation of Fujian Province of China (No.2021J01002).


%
%
\bibliographystyle{splncs04}
\bibliography{egbib}

\clearpage

\section*{Appendix \label{appendix}}

\section{More Ablation Studies}
\subsection{Effect of $\eta$ in Eq.~(5)}
\begin{figure}[h]
\centering
\includegraphics[width=1.0\textwidth]{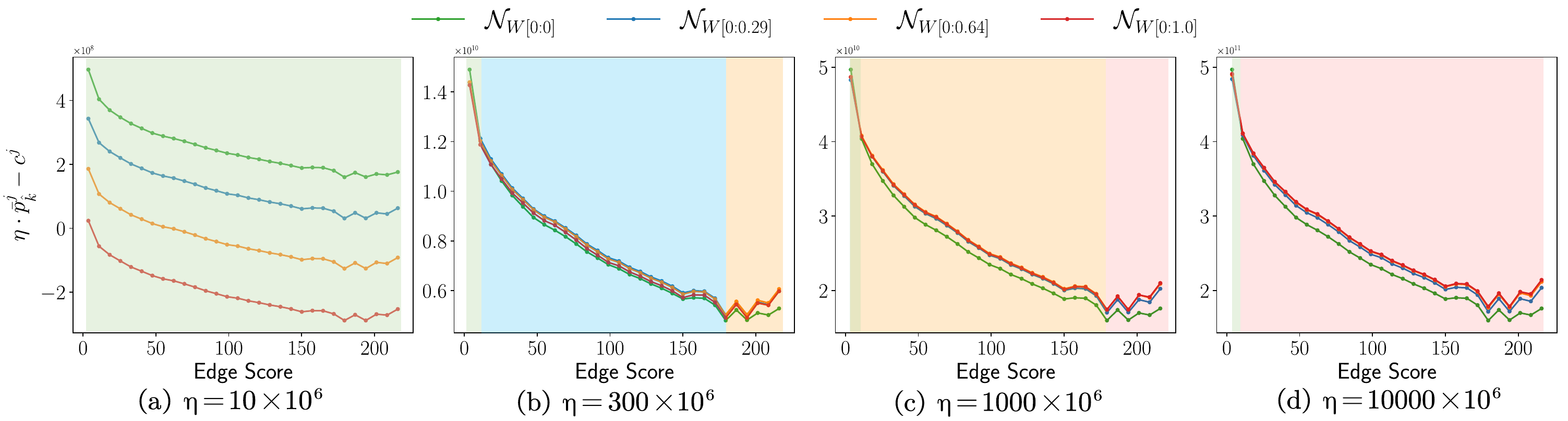}
\caption{The impact of $\eta$ in Eq.\,(5). 
The background color indicates the selected subnet or interpolation branch according to the edge score.
}
\label{fig:eta}
\end{figure}

Recall that in Eq.~(5), $\eta$ is a manually set hyper-parameter used to balance the numerical difference between the calculated cost and the PSNR estimate in the inference stage.
A larger $\eta$ leads to a preference to the subnet with a larger estimated PSNR $\bar{p}^j_{\hat{k}}$.
To better illustrate the process of subnet selection in Eq.\,(5), we plot the Edge-to-PSNR lookup tables of different subnets as well as the interpolation branch (denoted as $\mathcal{N}_{W[0:0]}$) using different $\eta$ in Fig.~\ref{fig:eta}.

As shown in Fig.~\ref{fig:eta}~(a), when the available computation is close to zero, we take a small value for $\eta$ and then ARM automatically selects the branch with the highest PSNR, \emph{i.e.}, interpolation branch ($\textcolor{green}{green}$ background), for all edge score patches to satisfy the computational constraint.
When there are more calculations available, we can set $\eta$ to a larger value (such as Fig.~\ref{fig:eta}~(b)(c)). In Fig.~\ref{fig:eta}~(b), for example, when the edge score of the patches is small (\emph{e.g.}, edge score $<$ 10), ARM will select the interpolation branch ($\textcolor{green}{green}$ background) for these patches; when the edge score is moderate (\emph{e.g.}, edge score $\in (10, 180)$), ARM will select $subnet_1$ ($\textcolor{blue}{blue}$ background) for these patches; For larger edge scores, $subnet_2$ ($\textcolor{yellow}{yellow}$ background) will be selected.
Finally, we set $\eta$ to a large value when there are many computation resources available. In this case, ARM selects the largest $subnet_3$ to super-resolution almost all patches. It is worth noting that for very ``easy" patches, ARM still chooses interpolation rather than always selecting the subnets, since for these patches, interpolation outperforms all sizes of subnets.

\begin{figure}
\centering
\includegraphics[width=0.95\textwidth]{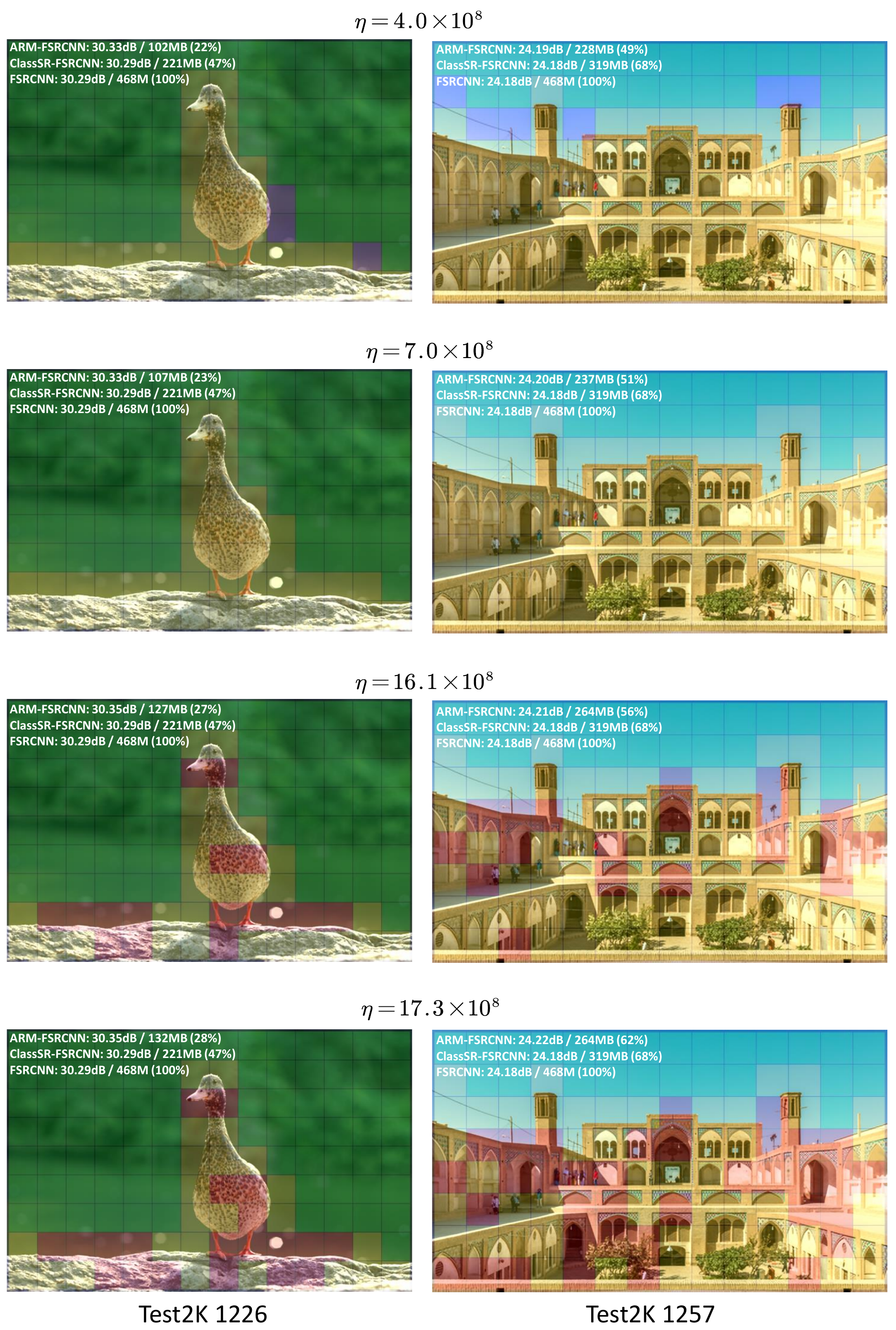}
\caption{Examples of super-resolution visualization of ARM-FSRCNN with different $\eta$. 
The {\color{green}green}, {\color{blue}blue}, {\color{yellow}yellow} and {\color{red}red} masks on the patch indicate that ARM uses interpolation, $subnet_1$, $subnet_2$ and $subnet_3$ for super-resolution of the patch, respectively.}
\label{fig:visual}
\end{figure}

We use some images from Test2K as examples to illustrate how ARM automatically adjusts the calculation based on the same pre-trained ARM supernet with different $\eta$ settings at inference time. 
The results are shown in Fig.~\ref{fig:visual}. As the $\eta$ increases, more and more patches are selected to use a larger subnet for super-resolution, thus gradually improving the PSNR.
It can also be seen that in Fig~\ref{fig:visual}, the ARM achieves better performance with fewer FLOPs than the backbone network and the previous SOTA dynamic SISR method~\cite{kong2021classsr}.

\subsection{Effect of $K$ subintervals}

\begin{wrapfigure}{r}{0.5\textwidth}
\centering
\includegraphics[width=0.5\textwidth]{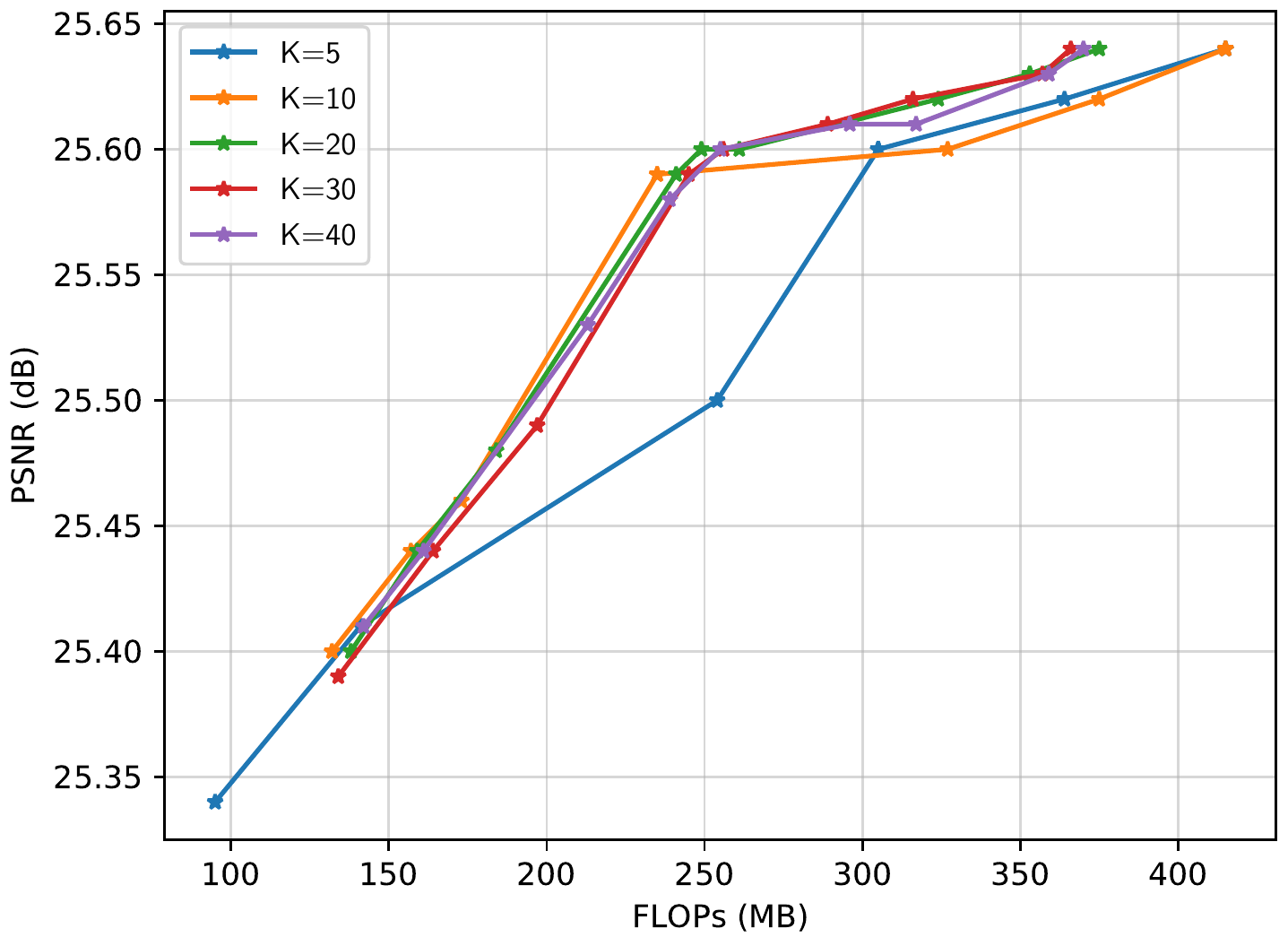}
\caption{Effect of different values of $K$.}
\label{fig:ab4}
\end{wrapfigure}

Our Edge-to-PSNR Lookup Tables are constructed by splitting the edge score interval into a total of $K$ subintervals and then averaging over all PSNR values within each interval as the estimated PSNR. Fig.\,\ref{fig:ab4} analyzes the impact of $K$.
For a small value of $K$, the estimated PSNR is loosely scattered, which causes an inaccurate PSNR estimation.
Increasing the value of $K$ leads to a more well-fitted edge-psnr mapping.
Though a larger $K$ may result in a better estimation, more parameters from the lookup tables are introduced.
In our experiment, we set $K=30$ across all the experiments for a better tradeoff and observe high-performing results as well.

\begin{table}[ht]
    \centering
    \caption{The results of latency (Lat.) of Test2K with FSRCNN as the backbone. The latency is the average of five trials.}
    \setlength{\tabcolsep}{0.6mm}{
    \begin{tabular}{l|c|c|ccc} 
    \toprule
    Model & FSRCNN & ClassSR & ARM-L & ARM-M & ARM-S \\ 
    \hline
    FLOPs/M & 468 (100\%) & 311 (66\%) & 366 (78\%) & 289 (62\%) & 245 (52\%) \\
    Lat./ms & 542.25 & 549.48 & 540.55 & 531.36 & 518.96 \\
    \bottomrule
    \end{tabular}
    }
    \label{tab2}
\end{table}

\section{The inference speed}

We measure the average inference time per image on Test2K dataset 
on a single core of \emph{Intel Xeon Platinum 8255C} CPU. The results are listed in Table~\ref{tab2}. Due to limited rebuttal time, no further speed optimization was conducted for hardware, so the actual speedup ratio was somewhat different from the theoretical speedup ratio. Despite, ARM still has some speed advantage over backbone and ClassSR.

\clearpage

\end{document}

%% file: tab_MRonFlickr2KTest2K.tex
\begin{table}[t]
\centering
\caption{The comparison of various methods and our ARM on F2K and Test2K.}
\label{tab:MRonFlickr2KTest2K}
\resizebox{.9\linewidth}{!}
{
\begin{tabular}{c|c|cc|cc} 
\toprule
Model & Params & F2K & FLOPs & Test2K & FLOPs \\ 
\hline
FSRCNN~\cite{dong2016accelerating} & 25K (100\%) & 27.91 (+0.00) & 468M (100\%) & 25.61 (+0.00) & 468M (100\%) \\
ClassSR~\cite{kong2021classsr} & \begin{tabular}[c]{@{}c@{}}113K (452\%)\\\end{tabular} & 27.93 (+0.02) & 297M (63\%) & 25.61 (+0.00) & 311M (66\%) \\ 
\hline
ARM-L & \multirow{3}{*}{25K (100\%)} & 27.98 (+0.07) & 380M (82\%) & 25.64 (+0.03) & 366M (78\%) \\
ARM-M &  & 27.91 (+0.00) & 276M (59\%) & 25.61 (+0.00) & 289M (62\%) \\
ARM-S &  & 27.65 (-0.26) & 184M (39\%) & 25.59 (-0.02) & 245M (52\%) \\ 
\hline\hline
CARN~\cite{ahn2018fast} & 295K (100\%) & 28.68 (+0.00) & 1.15G (100\%) & 25.95 (+0.00) & 1.15G (100\%) \\
ClassSR~\cite{kong2021classsr} & 645K (219\%) & 28.67 (-0.01) & 766M (65\%) & 26.01 (+0.06) & 841M (71\%) \\ 
\hline
ARM-L & \multirow{3}{*}{295K (100\%)} & 28.76 (+0.08) & 1046M (89\%) & 26.04 (+0.09) & 945M (80\%) \\
ARM-M &  & 28.68 (+0.00) & 819M (69\%) & 26.02 (+0.07) & 831M (71\%) \\
ARM-S &  & 28.57 (-0.11) & 676M (57\%) & 25.95 (+0.00) & 645M (55\%) \\ 
\hline\hline
SRResNet~\cite{ledig2017photo} & 1.5M (100\%) & 29.01 (+0.00) & 5.20G (100\%) & 26.19 (+0.00) & 5.20G (100\%) \\
ClassSR~\cite{kong2021classsr} & 3.1M (207\%) & 29.02 (+0.01) & 3.43G (66\%) & 26.20 (+0.01) & 3.62G (70\%) \\ 
\hline
ARM-L & \multirow{3}{*}{1.5M (100\%)} & 29.03 (+0.02) & 4.23G (81\%) & 26.21 (+0.02) & 4.00G (77\%) \\
ARM-M &  & 29.01 (+0.00) & 3.59G (69\%) & 26.20 (+0.01) & 3.48G (67\%) \\
ARM-S &  & 28.97 (-0.04) & 2.74G (53\%) & 26.18 (-0.01) & 2.87G (55\%) \\
\bottomrule
\end{tabular}
}
\vspace{-1.2em}
\end{table}

%% file: tab_MRonTest4KTest8K.tex
\begin{table}[t]
\centering
\caption{The comparison of various methods and our ARM on Test4K and Test8K.}
\label{tab:MRonTest4KTest8K}
\resizebox{.92\linewidth}{!}
{
\begin{tabular}{c|c|cc|cc} 
\toprule
Model & Params & Test4K & FLOPs & Test8K & FLOPs \\ 
\hline
FSRCNN~\cite{dong2016accelerating} & 25K (100\%) & 26.90 (+0.00) & 468M (100\%) & 32.66 (+0.00) & 468M (100\%) \\
ClassSR~\cite{kong2021classsr} & 113K (452\%) & 26.91 (+0.01) & 286M (61\%) & 32.73 (+0.07) & 238M (51\%) \\ 
\hline
ARM-L & \multirow{3}{*}{25K (100\%)} & 26.93 (+0.03) & 341M (73\%) & 32.75 (+0.09) & 290M (62\%) \\
ARM-M &  & 26.90 (+0.00) & 282M (60\%) & 32.73 (+0.07) & 249M (53\%) \\
ARM-S &  & 26.87 (-0.03) & 230M (50\%) & 32.66 (+0.00) & 187M (40\%) \\ 
\hline\hline
CARN~\cite{ahn2018fast} & 295K (100\%) & 27.34 (+0.00) & 1.15G (100\%) & 33.18 (+0.00) & 1.15G (100\%) \\
ClassSR~\cite{kong2021classsr} & 645K (219\%) & 27.42 (+0.08) & 742M (64\%) & 33.24 (+0.06) & 608M (53\%) \\ 
\hline
ARM-L & \multirow{3}{*}{295K (100\%)} & 27.45 (+0.11) & 825M (70\%) & 33.31 (+0.13) & 784M (66\%) \\
ARM-M &  & 27.42 (+0.08) & 743M (64\%) & 32.27 (+0.09) & 612M (53\%) \\
ARM-S &  & 27.34 (+0.00) & 593M (50\%) & 33.18 (+0.00) & 489M (42\%) \\ 
\hline\hline
SRResNet~\cite{ledig2017photo} & 1.5M (100\%) & 27.65 (+0.00) & 5.20G (100\%) & 33.50 (+0.00) & 5.20G (100\%) \\
ClassSR~\cite{kong2021classsr} & 3.1M (207\%) & 27.66 (+0.01) & 3.30G (63\%) & 33.50 (+0.00) & 2.70G (52\%) \\ 
\hline
ARM-L & \multirow{3}{*}{1.5M (100\%)} & 27.66 (+0.01) & 3.41G (66\%) & 33.52 (+0.02) & 3.24G (62\%) \\
ARM-M &  & 27.65 (+0.00) & 3.24G (62\%) & 33.50 (+0.00) & 2.47G (48\%) \\
ARM-S &  & 27.63 (-0.02) & 2.77G (53\%) & 33.46 (-0.04) & 1.83G (35\%) \\
\bottomrule
\end{tabular}
}
\end{table}